\documentclass{article}

\usepackage{microtype}
\usepackage{graphicx}
\usepackage{subfigure}
\usepackage{booktabs} 
\usepackage{amsfonts}
\usepackage{amsmath}

\usepackage{hyperref}

\usepackage[accepted]{icml2019}

\icmltitlerunning{Supervise Thyself}

\begin{document}

\twocolumn[
\icmltitle{Supervise Thyself: Examining Self-Supervised Representations in Interactive Environments}



\icmlsetsymbol{equal}{*}

\begin{icmlauthorlist}
\icmlauthor{Evan Racah}{mila}
\icmlauthor{Chris Pal}{mila,poly,elem}
\end{icmlauthorlist}

\icmlaffiliation{mila}{Mila, Universit\'e de Montr\'eal}
\icmlaffiliation{poly}{\'Ecole Polytechnique de Montr\'eal}
\icmlaffiliation{elem}{Element AI, Montr\'eal, QC, Canada }

\icmlcorrespondingauthor{Evan Racah}{ejracah@gmail.com}
\icmlcorrespondingauthor{Chris Pal}{chris.j.pal@gmail.com}

\icmlkeywords{Machine Learning, ICML}

\vskip 0.3in
]



\printAffiliationsAndNotice{}  

\begin{abstract}
Self-supervised methods, wherein an agent learns representations solely by observing the results of its actions, become crucial in environments which do not provide a dense reward signal or have labels.
In most cases, such methods are used for pretraining or auxiliary tasks for "downstream" tasks, such as control, exploration, or imitation learning.
However, it is not clear which method's representations best capture meaningful features of the environment, and which are best suited for which types of environments.
We present a small-scale study of self-supervised methods on two visual environments: Flappy Bird and \textit{Sonic The Hedgehog\textsuperscript{TM}}.
In particular, we quantitatively evaluate the representations learned from these tasks in two contexts: a) the extent to which the representations capture true state information of the agent and b) how generalizable these representations are to novel situations, like new levels and textures. Lastly, we evaluate these self-supervised features by visualizing which parts of the environment they focus on. Our results show that the utility of the representations is highly dependent on the visuals and dynamics of the  environment. 
\end{abstract}

\section{Introduction}
Self-Supervised methods have emerged as powerful methods for pretraining to learn more general representations for complicated downstream tasks in vision \cite{misra2016shuffle,fernando2015modeling,fernando2017self,wei2018learning,vondrick2018tracking,jayaraman2015learning,agrawal2015learning, pathak2017learning, wang2015unsupervised} and NLP \cite{peters2018deep,subramanian2018learning,mikolov2013efficient,conneau2018senteval}. In interactive environments, they have begun to receive more attention due their ability to learn general features of important parts of the environment without any extrinsic reward or labels  \cite{lecun2018selfsup}. Specifically, self-supervised methods has been used as auxiliary tasks to help shape the features or add signal to sparse reward problems \cite{mirowski2016learning,jaderberg2016reinforcement,shelhamer2016loss}. 
They also have been used in unsupervised pretraining for control problems, \cite{ha2018world}. Moreover, they have been used in imitation learning to push expert demonstrations and agent observations into a shared feature space  \cite{aytar2018playing,sermanet2017time}. Lastly, they have been used in intrinsic reward exploration to learn a representation well-suited for doing surprisal-based prediction in feature space \cite{pathak2017curiosity,burda2018large}. In each of these cases, the desired feature space learned with self-supervised methods should capture the agent, objects, and other features of interest, be easy to predict, and generalize to unseen views or appearances. However, existing evaluations of these methods do not really shed light on whether the representations learned by these self-supervised methods really robustly capture these things. Instead, these evaluations only evaluate the utility of these methods on the particular downstream task under study. While these types of tasks have been studied theoretically \cite{hyvarinen2018nonlinear,arora2019theoretical}, they have not really been examined empirically in depth. As such, in this paper we examine a few self-supervised tasks where we specifically try to characterize the extent to which the learned features capture the state of the agent and important objects. Specifically, we measure how well the features: capture the agent and object positions and
generalize to unseen environments, and lastly, we  qualitatively measure what each self-supervised method is focusing on in the environment. 
We pick Flappy Bird and \textit{Sonic The Hedgehog\textsuperscript{TM}} because they represent simple and complex games respectively in terms of graphics and dynamics. Also, one can change the level and colors of each to make an "unseen" environment to test generalizability.  
\footnote{https://github.com/eracah/supervise-thyself}
\section{The Self-Supervised Methods We Explore}
\begin{figure*}[ht!]
\vskip 0.2in
\begin{center}
\label{arch}
\includegraphics[scale=0.6]{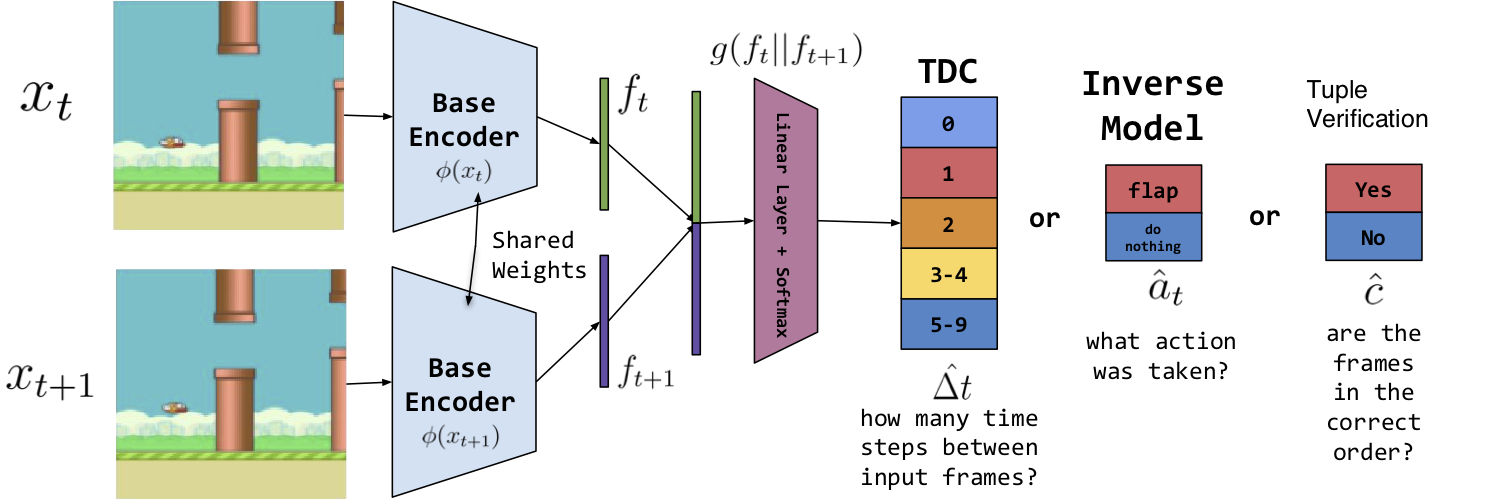}
\caption{ \textbf{General architecture for self-supervised embedding}. Shown for Flappy Bird. Two or three frames are each input to the base encoder then the outputs from the encoder, $\phi(x)$ are concatenated and passed to a linear softmax layer that classifies either a) "how many time steps are between a pair of frames?" for the TDC model \cite{aytar2018playing}, b) "what action was taken to go from the first frame to second?" for the inverse model \cite{agrawal2016learning}, or c) "are a triplet of frames in the correct chronological order?" for the tuple verification model \cite{misra2016shuffle}  }
\end{center}
\vskip -0.2in
\end{figure*}

We explore four different approaches for self-supervision in interactive environments: VAE \cite{kingma2013auto}, temporal distance classification (TDC)\cite{aytar2018playing} , tuple verification \cite{misra2016shuffle}, and inverse model \cite{agrawal2016learning,jayaraman2015learning,pathak2017curiosity}. We also use a randomly initialized CNN as a baseline.
All self-supervised models in this study use a \textit{base encoder}, $\phi$ which is a four layer convolutional neural network, similar in architecture to the encoder used in the VAE in \cite{ha2018world}. The encoder takes as input a single frame in pixel space, $x$ and outputs $f = \phi(x)$, where $f \in \mathbb{R}^{32}$. Depending on the self-supervised task, 
certain heads, $g$ are placed on top of the encoder, like a deconvolutional decoder or a linear classifier, that take z or multiple concatenated z's as input. See figure \ref{arch} and the appendix for more details.

\section{Experiments/Results}
\subsection{Datasets/Environments}
We collect the data for Flappy Bird \cite{tasfi2016PLE} and \textit{Sonic The Hedgehog\textsuperscript{TM}}, the \texttt{GreenHillZone} Act 1 level \cite{nichol2018gotta}  by deploying a random agent to collect 10,000 frames. At train time we randomly select frames from these 10,000 to form each batch. For the generalization experiments in section \ref{gen-new}, we use what we call \texttt{FlappyBirdNight}, whereby we change the background, the color of the bird and the color of the pipes. For generalization in Sonic, we use \texttt{GreenHillZone} Act 2. We describe the dataset collection in more detail in the appendix
\subsection{Extracting Position Information }
\label{ext-state-inf}
To show whether the self-supervised methods are capable of encoding the position of important objects, we probe the representations learned from these methods with a linear classifier trained to classify the agent position (bucketed to 16).
The results of this experiment are displayed in table \ref{pos-inf} We find features from the inverse model are the most discriminative for detecting the position of the bird, likely because the inverse model focuses on what parts of the environment it can predictably control. 
We find features from TDC are the best for localizing the pipe. This is likely the case because TDC focuses on what parts of the environment are most discriminative for separating frames in time and in Flappy Bird, the background and bird stay in one place and the pipes move to the left to simulate the bird moving to the right. Tuple verification features are good for both objects, the pipe and the bird, likely because the position of the pipe relative to the bird is a very important temporal signal, which is discriminative to whether the frames are in order or not. The VAE does not do much better than random features for the small-sized bird, but very respectably for the large pipe, likely due to VAE's preference to capture larger global structure that more contributes to the reconstruction loss. \\
Sonic, on the other hand, has much more complex dynamics and graphics than Flappy Bird. As a result the classification performance is not as strong. For example, the inverse model does much worse at capturing the position of Sonic. This is likely due to the more inconsistent response of Sonic to action commands. For example, when Sonic is already in the air jumping, the right command has no effect. The ambiguity to which action was called for what pairs of frames, likely causes the inverse model to do bad at its task and thus not learn good features. Moreover, the frame moves up in response to Sonic jumping, so Sonic's exact pixels are not the only thing that change in response to jump, making it tougher for the inverse model to focus in on Sonic. Moreover, sequence verification methods like TDC and tuple verification are also tripped up by Sonic. This is most likely because even though Sonic moves to the right fairly consistently, the background moves in the x position not Sonic. Normally, that would be no problem for TDC and tuple verification, like in Flappy Bird. However, there is no consistent landmark in the background for these methods to use like the pipes in Flappy Bird. VAEs also do worse than they do at Flappy Bird. However, they do relatively better than any other self-supervised methods. Likely, this is because they are not affected by weird dynamics of the environment.


\subsection{Generalizing to New Levels and Colors}\label{gen-new}
We can also show how well these features generalize to new situations. Theoretically, if the features are truly, robustly capturing objects of interest, changing the level or the colors of the environment, should not affect a linear classifier's ability to localize objects of interest.  We test this out by looking at zero-shot linear probe accuracy with the background, pipe, and bird colors changed for Flappy Bird and on a new level for Sonic.We see these results in table \ref{gen-pos}.
Unsurprisingly, we find the performance decreases for all self-supervised methods in Flappy Bird. Surprisingly, the features from TDC generalize better than the inverse model for classifiying bird's location. Potentially, this is because the color of the bird changes and the features from the inverse model are more specific to the exact appearance of the bird from the setup it was trained on. TDC features, on the hand, may encode the bird based on where it is relative to the pipes, and less so on exactly how it looks and for the same reason, TDC features are able to capture the pipes, despite their different color. The VAE features' performance unsurprisingly drops for both objects, as the global structure that they learn to encode completely changes with the new colors in the \texttt{FlappyBirdNight} setup.
\subsection{Qualititative Inspection of Feature Maps}
\begin{figure*}[ht!]
\vskip 0.2in
\begin{center}
\includegraphics[width=0.15\linewidth]{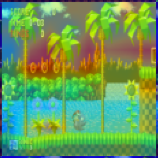}
 \includegraphics[width=0.15\linewidth]{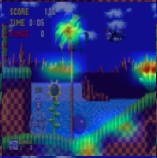}
 \includegraphics[width=0.15\linewidth]{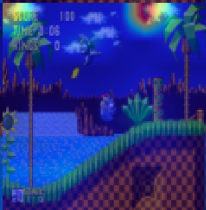}
  \includegraphics[width=0.15\linewidth]{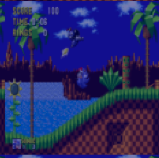}
 \includegraphics[width=0.15\linewidth]{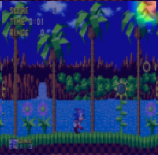} 
\caption{Sonic feature maps from the last conv layer of the encoder superimposed on top of the frames they are a function of for from left: random CNN, VAE, inverse Model, tuple verification, and temporal distance classification. Red is high values, blue are low values}
\label{sonic-fmaps}
\end{center}
\vskip -0.2in
\end{figure*}

We show qualitative inspection by superimposing a frame’s feature map on top of the frame itself. We pick the most compelling feature map for each frame, which we show in figures \ref{flappy-fmaps} and \ref{sonic-fmaps}. Confirming our hypothesis from \ref{ext-state-inf}, we see for Flappy Bird, the inverse model feature map focuses on the bird and the TDC feature map focuses on the pipe. Interestingly enough, tuple verification keys in on the top half of the pipe and the VAE activates on everything in the frame, but the pipes.
For Sonic, things are not as clean and interpretable. As we see in figure  \ref{sonic-fmaps}, the inverse model feature map and the TDC one focus in on a cloud, perhaps mistaking it for Sonic, and the tuple verification map keys in on nothing at all. The VAE feature map, unsurprisingly, activates on important, ubiquitous objects for reconstructing the frame, like the tree and the bush. None of these representative feature maps key in on Sonic himself, which agrees with the poor quantitative classification accuracy results in table \ref{pos-inf}.
\begin{figure}[ht!]
\vskip 0.001in
\begin{center}
\includegraphics[width=0.95\linewidth]{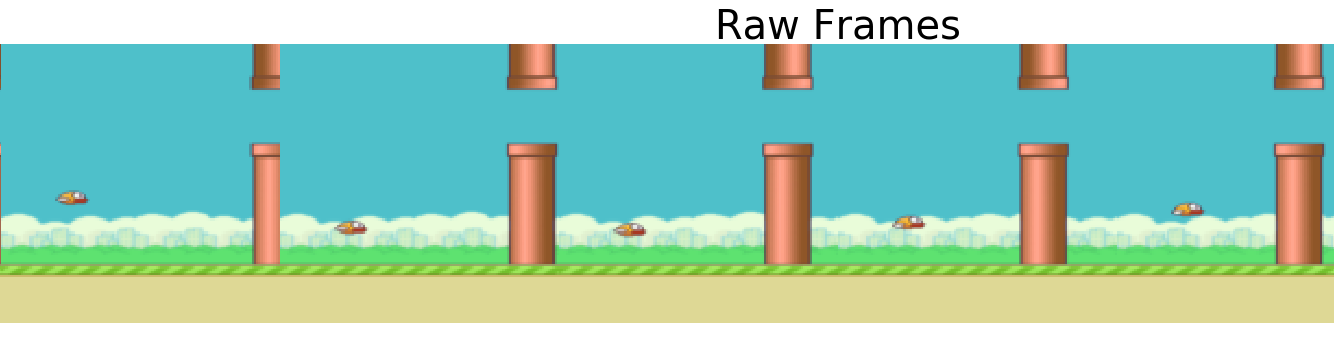}
\includegraphics[width=0.95\linewidth]{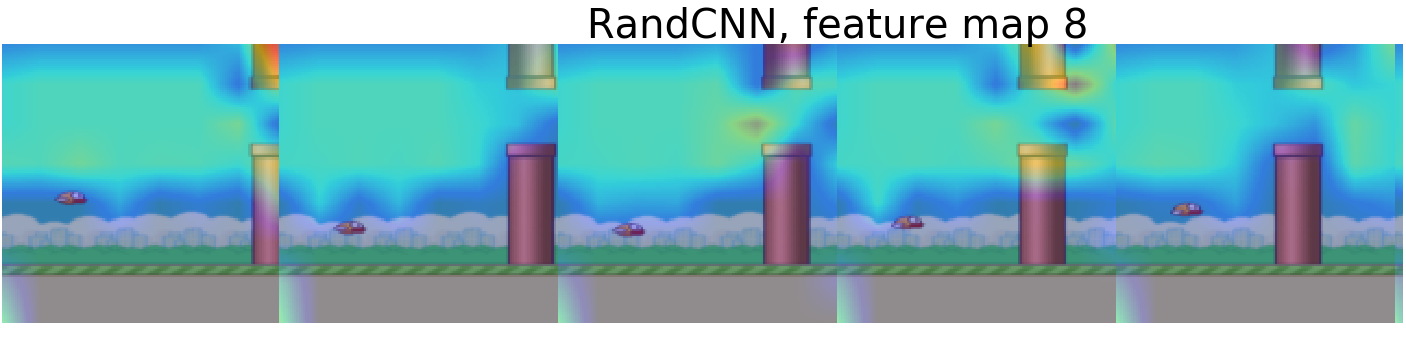}
\includegraphics[width=0.95\linewidth]{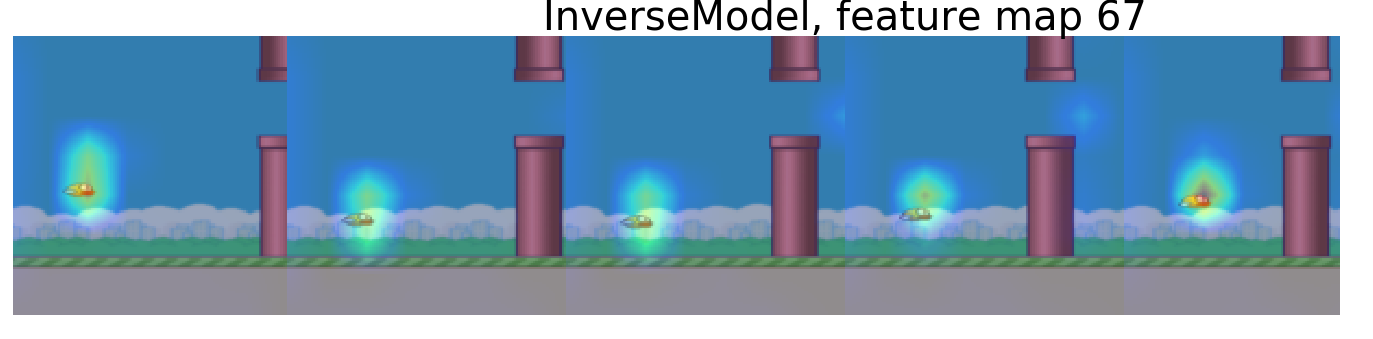}
\includegraphics[width=0.95\linewidth]{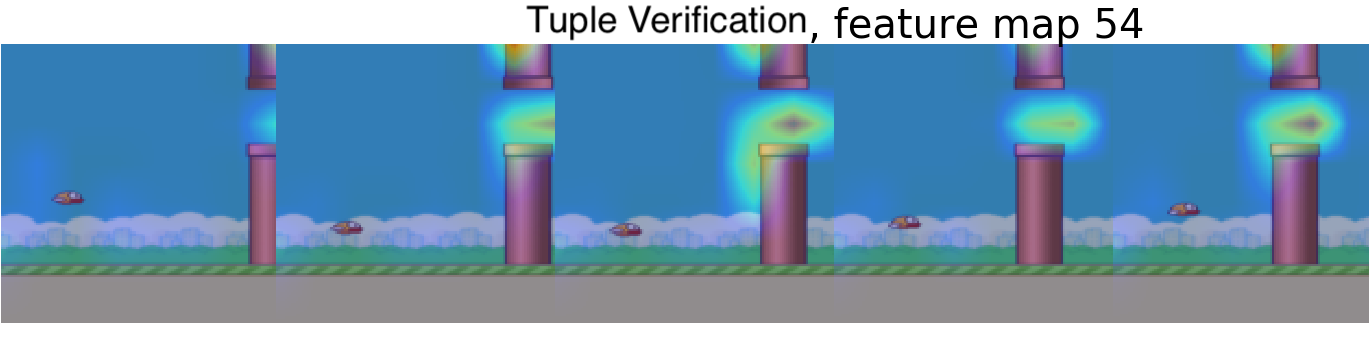}
\includegraphics[width=0.95\linewidth]{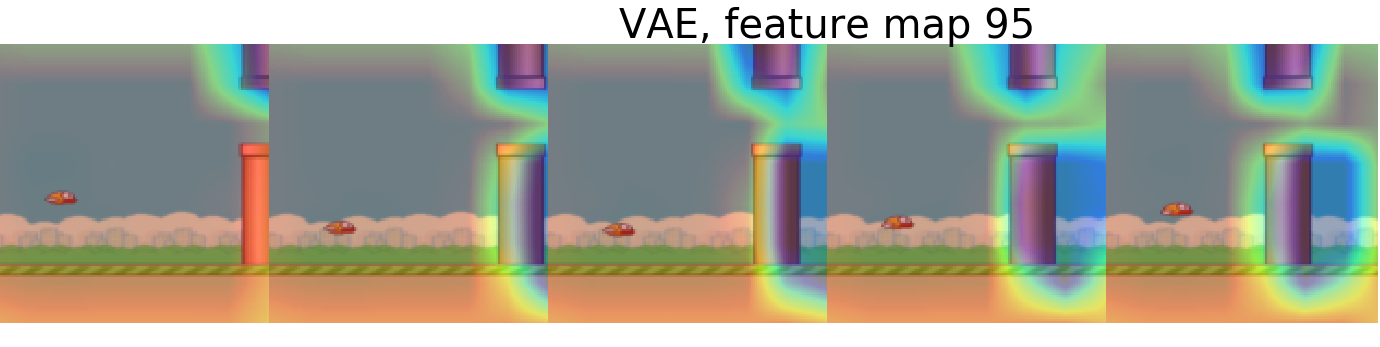}
\includegraphics[width=0.95\linewidth]{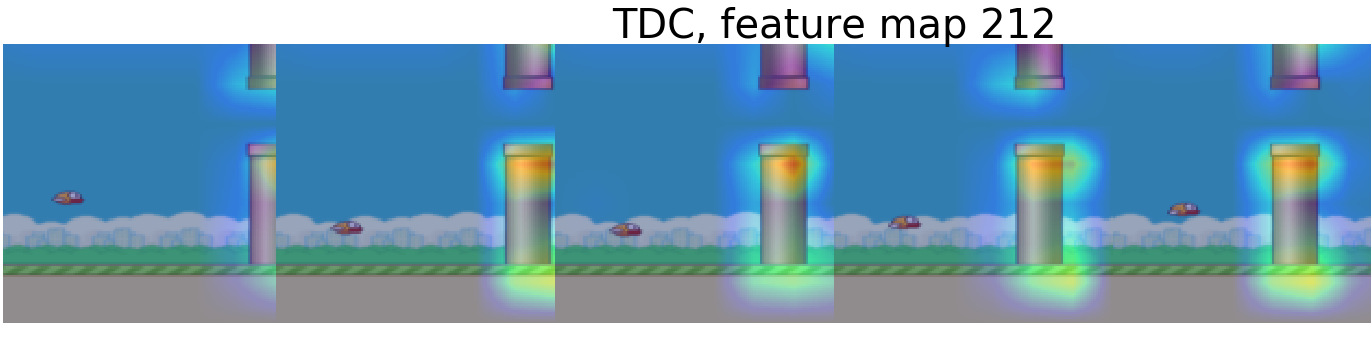}

\caption{\textbf{Qualititative Inspection of Feature Maps} Flappy Bird feature maps from the last conv layer of the encoder superimposed on top of a sequence of frames they are a function of. Red pixels are high values, blue are low values}
\label{flappy-fmaps}
\end{center}
\vskip -0.2in
\end{figure}

\begin{table*}[t]
\caption{\textbf{Extracting Position Information} We train a linear classifier on top of feature spaces from each method and measure the classification accuracy for the various position of various objects in the game, y position of the bird and x position of the first pipe for Flappy Bird and y position of Sonic in Sonic }
\label{pos-inf}
\vskip 0.15in
\begin{center}
\begin{small}
\begin{sc}
\begin{tabular}{@{}lcccr@{}}
\toprule
\textbf{Method}      &  \multicolumn{2}{c}{Flappy Bird} & Sonic  \\
& 	Bird Y Pos Acc (\%)  & Pipe X Pos Acc(\%) & Y Pos (\%) \\
\midrule
    RandCNN    & 35.57          & 52.13          & 23.88 \\
    VAE        & 36.20          & 76.52          & \textbf{53.82} \\        
    Inv Model  & \textbf{91.67} & 81.64          & 25.83\\
    tuple verification        & 75.41          & 87.08          & 11.2 \\
    TDC        & 56.06          & \textbf{92.36} & 10.42  \\        

\bottomrule
\end{tabular}
\end{sc}
\end{small}
\end{center}
\vskip -0.1in
\end{table*}


\begin{table*}[t]
\caption{\textbf{Generalizing Extracting Position to New Levels and Colors} We see how well the trained linear classifiers do in a zero shot transfer to new colors for Flappy Bird and a new level for Sonic}
\label{gen-pos}
\vskip 0.15in
\begin{center}
\begin{small}
\begin{sc}
\begin{tabular}{lcccr}
\toprule
\textbf{Method}      &  \multicolumn{2}{c}{\texttt{FlappyBirdNight}} & Sonic \texttt{GreenHillZone} Act 2 \\
& 	                Bird Y Pos Acc (\%)  & Pipe X Pos Acc(\%) & Y Pos (\%) \\
\midrule
    RandCNN           & 34.03           & 7.41               & 28.31                 \\
    VAE               & 2.53            & 2.0                & \textbf{34.57}                \\   
    Inv Model         & 36.5            & 8.90               &  27.28              \\
    tuple verification               & 1.88            & 1.74               &  14.4              \\
    TDC               & \textbf{46.9}   & \textbf{16.22}     &  13.59     \\        

\bottomrule
\end{tabular}
\end{sc}
\end{small}
\end{center}
\vskip -0.1in
\end{table*}

\section{Related Work}
This paper is not the first paper to quantitatively and qualitatively compare features from self-supervised methods in interactive environments. \cite{burda2018large} compare the feature spaces learned from VAE's, Inverse Models, and Random CNN's and raw pixels (but not sequence verification methods) across a large variety of games. They even measure the generalization of these feature spaces to new, unseen environments. However, all their evaluations are in the context of how well an agent explores its environment with this feature space using extrinsic rewards and other measures of exploration. Additionally, \cite{shelhamer2016loss} studies VAE's, Inverse Models, and a sequence verification task in RL environments, but they evaluate these self-supervised methods as auxiliary tasks paired with a traditional extrinsic reward policy gradient algorithm, A3C \cite{mnih2016asynchronous}, using empirical returns from extrinsic rewards as a measure of utility of each feature space. Lastly, trying to infer the position of salient objects has been explored a lot in robotics in a field called state inference \cite{jonschkowski2015learning,jonschkowski2017pves}. Moreover, \cite{raffin2018s,lesort2018state} look into using some self-supervised tasks for state inference, but they mostly measure performance on control tasks; they do not measure direct correlation or classification accuracy of the features to the true position of the object.
 
\section{Discussion}
We have shown comparing methods on interactive environments reveals intriguing things about the self-supervised methods as well as the environments themselves.
Particularly, we expose various traits of environments that some self-supervised tasks can take advantage of and others cannot. For example, inverse models are very good at localizing what they can control even it is small, but only when the dynamics are simple and predictable and the appearance of the agent itself is consistent. Temporal distance classifiers are very good at capturing things that move very predictably in time. Tuple verification encoders are good at capturing small and large objects in environments with pretty consistent graphics and dynamics. VAE's learn good features when the objects are big and repeatably show up in the scene with consistent appearance.
\subsection{Future Work} 
The very different behaviors of each method depending on the traits of the environment warrants further study covering more environments with more diverse appearances and dynamics, as well as a wider range of self-supervised methods. In addition, the fact that some methods excel at capturing or generalizing better than others depending on the environment motivates exploring potentially combining these methods by having a shared encoder body with multiple self-supervised heads. We hope that this study can open the door to more extensive, rigorous approaches for studying the capability of self-supervised methods and that its results can inspire new methods that learn even better features.

\bibliography{my_bib}
\bibliographystyle{icml2019}
\newpage
\newpage
\appendix
\icmltitle{Appendix}
\section{More Details on The Self-Supervised Methods We Explore}
Code available at https://github.com/eracah/self-supervised-survey
Most of the self-supervised tasks work like so:
\begin{equation}
\hat o = g([f_i,...,f_{j}]) \\
\end{equation}
where $[f_i,...,f_{j}]$ is a concatenation of anywhere from 1 to 3 embeddings depending on the self-suepervised method, where $g$ is a deconvolutional decoder for VAE's and a linear classifier for the other self-supervised methods and: \\
\begin{equation}
f_t,  f_{t+1}= \phi(x_t), \phi(x_{t+1})
\end{equation}
The meaning of $\hat o$ varies depending on the task
\textbf{Random CNN}: We use a randomly initialized CNN as a baseline method. In this case, it is an untrained base encoder, $\phi$, with randomly initialized weights that are not updated. Random CNN's have been used with varying degrees of success in \cite{burda2018large,anon2018exploranddistill}

\textbf{VAE}: VAE's \cite{kingma2013auto} are latent variable models that maximize a lower bound of the data likelihood, $p(x)$ by approximating the posterior $p(z|x)$, with a parametric distribution $q(z|x)$ and a prior $p(z)$. VAE's also include a decoder $p(x|z)$, which reconstructs the input x by mapping samples of $z$ back to pixel space.In our setup $q(z|x)$ is parametrized by a gaussian with mean $\phi(x)$ and the variance is parametrized by a separate fully connected layer on top of the penultimate layer of the base encoder. Also, we use a deconvolutional network, $ \hat x = g(z)$ to parametrize $p(x|z)$. The VAE is trained by minimizing the KL divergence between the prior, $p(z)$, which we often pick to be an isoptropic guassian, and the approximate posterior,$q(z|x)$, while also minimizing the negative log-likelihood of $p(x|z)$, like so:
\begin{equation}
KL(q(z|x)||p(z)) - log p(x|z)
\end{equation}
The idea is that if we learn close to factorized latent variables that encode enough information to reliably reconstruct the frame, they will capture important structure of the image, like objects.

\textbf{Temporal Distance Classification (TDC)}: Temporal Distance Classification (TDC) is a self-supervised method introduced by \cite{aytar2018playing}, similar to \cite{hyvarinen2016unsupervised}, where the network learns the relative distance in time between frames.  In TDC, we do the following:
\begin{equation}
\hat \Delta t = g([f_t,f_{t+\Delta t}])
\end{equation}
 $[f_t,f_{t+\Delta t}]$ is the concatenation of the embeddings of frames $x_t$ and $x_{t+\Delta t}$,
and where $\Delta t$ is sampled from a set of intervals, $\Delta t \in D$ \\
\begin{equation}
D = \{[0],[1],[2],[3-4],[5-9]\} \\
\end{equation}
where $g$ is a k-way linear classifier, where $k = |D|$ that classifies which time interval separates the two input frames.
The idea is that in order to do well at the task, it must learn the features of the input that change over time, which often corresponds to objects \cite{aytar2018playing}.

\textbf{Tuple Verification}: Tuple Verification \cite{misra2016shuffle} is an instance of a temporal order or dynamics verification task, where the network must figure out if a sequence of frames is in order or not. The method works as such: five chronologically ordered, evenly-spaced (they don't have to be consecutive) frames are chosen from a rollout: 
\begin{equation}
\{x_a, x_b, x_c, x_d, x_e: a<b<c<d<e\}
\end{equation}
In this paper we use an evenly spaced sampling of 5 frames from a sequence of 10 consecutive frames. 
A binary classification problem is created by shuffling the frames to create negative examples. A tuple of frames in the order $x_b, x_c, x_d$ is a positive example, whereas the tuples $(x_b, x_e, x_d)$ and $(x_b, x_a, x_d)$ are negative examples. We ensure that there is a $2:1:1$ ratio between positive samples and the two types of negative samples. 
The three frames are each encoded with the base encoder, concatenated and passed to a linear classifier: \\
\begin{equation}
\hat c = g([f_b, f_c,f_d]), \hat c = g([f_b, f_e,f_d]), \hat c =  g([f_b, f_a,f_d])
\end{equation} 
where $g$ is linear binary classifier the predicts whether the frames are in order or not.
Being successful at this task requires knowing how objects transform and move over time, which requires encoding of features corresponding to the appearance and location of objects \cite{misra2016shuffle}.

\textbf{Inverse Model}: The inverse model \cite{agrawal2016learning,jayaraman2015learning,pathak2017curiosity} works by taking two consecutive frames from a rollout from an agent, then classifying which action was taken to transition from one frame to other. The model works like so:
\begin{equation}
\hat a_{t+1} = g([f_t, f_{t+1}])
\end{equation}
where $g$ is a k-way linear classifier, where k is the size of thr action space, the number of possible actions the agent can take.
The idea is that in order to reason about which action was taken, the network must learn to focus on parts of the environment are controllable and affect the agent \cite{choi2018contingency}. This should then result in the network learning features that capture the agent's state and location as well potential obstacles to the agent.

\section{More on Dataset Collection}
We edit the action space of Sonic to be just the two actions: ["Right"] and ["Down", "B"]. This ensures the random agent will get pretty far in the level and actually collect a good diversity of frames. For both games, we resize the frames to 128 x 128. All ground truth position information (y position of bird, x position of pipe, y position of Sonic) is pulled from the gym API of these games and discretized to 16 buckets and represents the relative position of these objects in the frame, not the absolute position in the game. We purposely do not choose the x position of the bird or Sonic because in most frames of the game, the x position is relatively constant, while the background moves.

\section{Prediction in Feature Space} 
\begin{figure}[ht!]
\vskip 0.2in
\begin{center}
\includegraphics[scale=0.55]{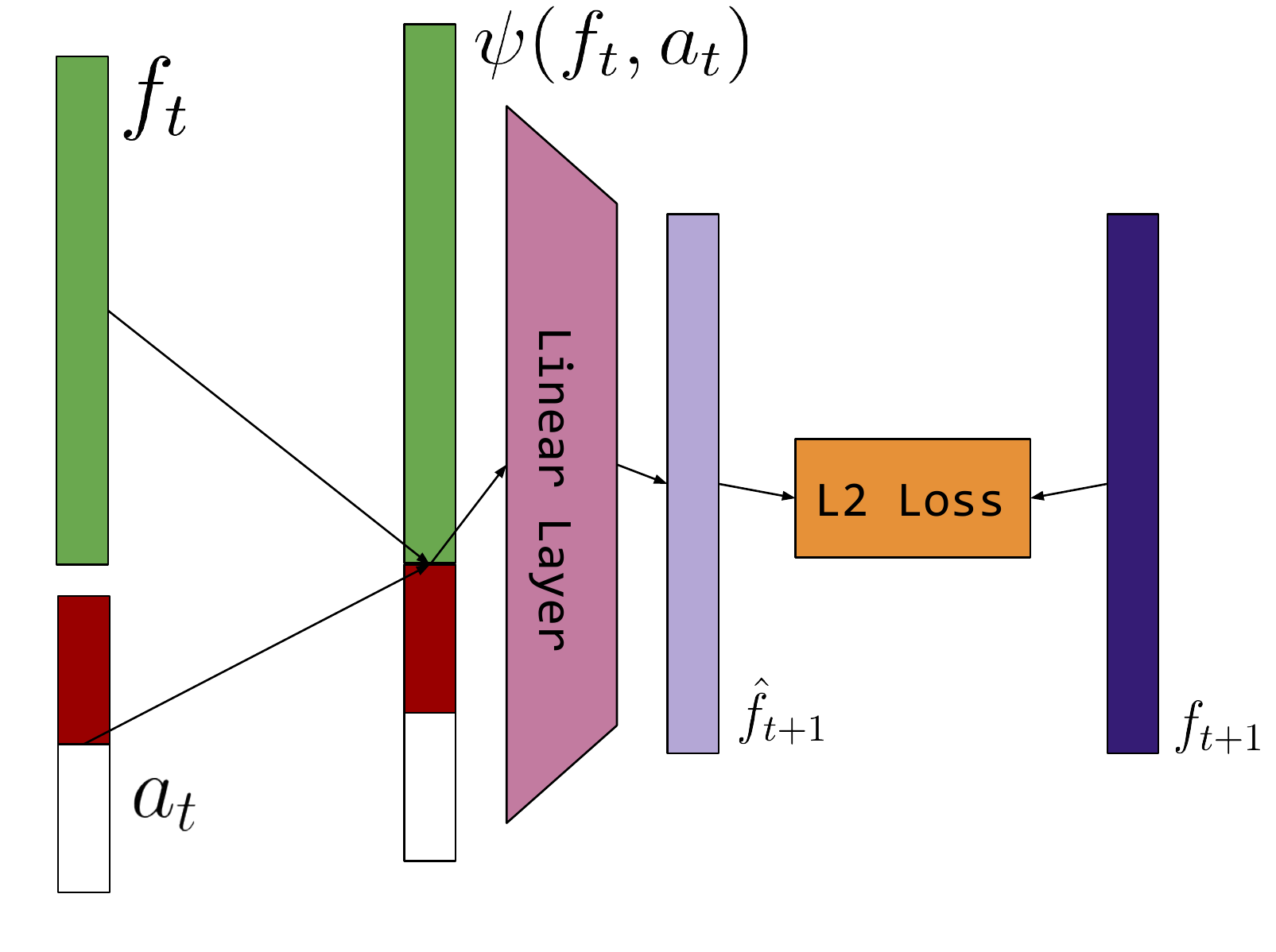}
\caption{\textbf{Predicting in Feature Space}: Architecture for predicting in feature space: an embedding at time step is concatenated with the action at time t and put through a linear layer to get the predicted embedding at time step t+1}
\label{pred-fig}
\end{center}
\vskip -0.2in
\end{figure}

To measure how predictive features are we train a one-step forward linear model, $\psi$ that takes in the features at time step t, $f_{t} = \phi(x_t)$ and the action, $a_t$, and predicts the next step's features, $ \hat f_{t+1}$, where the true future features are $f_{t+1} = \phi(x_{t+1})$ as seen in figure \ref{pred-fig}. \\
\begin{equation}
\hat f_{t+1} = \psi(f_{t},a_t)
\end{equation}
The loss is then the mean squared error between the true and predicted features: 
\begin{equation}
L_f = \frac{1}{m}|| f_{t+1} - \hat f_{t+1} ||_{2}^{2}
\end{equation} 
where m is number of features in the mebddings, which in our case is 32. \\
Then at test time, we use the forward model to iteratively get a rollout of ten time steps worth of features: \\

\begin{align} 
\hat f_{t+1}&=\psi(f_{t},a_t), \\
\hat f_{t+2} &=\psi(\hat f_{t+1},a_{t+1}),\\
\hat f_{t+3} &=\psi(\hat f_{t+2},a_{t+2}), \\
... \\
\hat f_{t+9} &=\psi(\hat f_{t+8},a_{t+8})
\end{align} \\

We then take the first principal component of each predicted feature and we look at Spearman correlation of the principal components with the true ground truth state information. 
\subsection{Results} We take a closer look at how "predictive" the features for each method are in table \ref{pred-pca}. Surprisingly, the predicted features better capture this state information than the true features for many self-supervised tasks. This might be because prediction is easier for simple objects than it is other (potentially more trivial) factors of variation, so the task of prediction slowly changes the feature space to capture less other factors of variation and more the ones for objects. In addition, in learning a dynamic task like prediction the forward model learns about the things that move. This could also be due what \cite{agrawal2016learning} refers to as regularizing the feature space with a forward model.  It is worth noting that the predicted inverse model features are very predictive, as they retain their stellar ability to capture the true agent position.

\begin{table*}[t]
\caption{\textbf{Evaluating Prediction in Feature Space }. We take the first principal component of the predicted feature vectors and compute the Spearman's rank correlation coefficient with the true values: y position of the bird and x position of the first pipe for Flappy Bird and y position of the Sonic character in Sonic}
\label{pred-pca}
\vskip 0.15in
\begin{center}
\begin{small}
\begin{sc}
\begin{tabular}{lcccccccr}
\toprule
&  \multicolumn{4}{c}{Flappy Bird} & \multicolumn{2}{c}{Sonic}  \\
\textbf{Method}   & 	\multicolumn{2}{c}{Bird Y Pos}  & \multicolumn{2}{c} {Pipe X Pos}  & \multicolumn{2}{c}{Sonic Y Pos} \\
& True & Predicted & True & Predicted & True & Predicted \\
\midrule
    RandCNN      & -0.01 &   0.066  & -0.078    & 0.0035         & \textbf{0.66}                  &    0.0044 \\
    VAE          &  -0.17 &   -0.21   & -0.32  & -0.59            &  0.52                   &    \textbf{0.011} \\
    InvModel     &  \textbf{0.88} &   \textbf{0.87} &  -0.092 & 0.04  &  0.04                 &   -0.0059\\     
    SNL          &   0.0   &   0.23  & \textbf{-0.82} & 0.0027 &  0.0016           &   0.0063\\
    TDC          &   -0.23 & -0.10          &  0.34 & \textbf{-0.71 }  &   -0.21           &   0.0072\\
\bottomrule
\end{tabular}
\end{sc}
\end{small}
\end{center}
\vskip -0.1in
\end{table*}

\begin{figure*}[ht!]
\vskip 0.2in
\begin{center}
  
\includegraphics[width=0.95\linewidth]{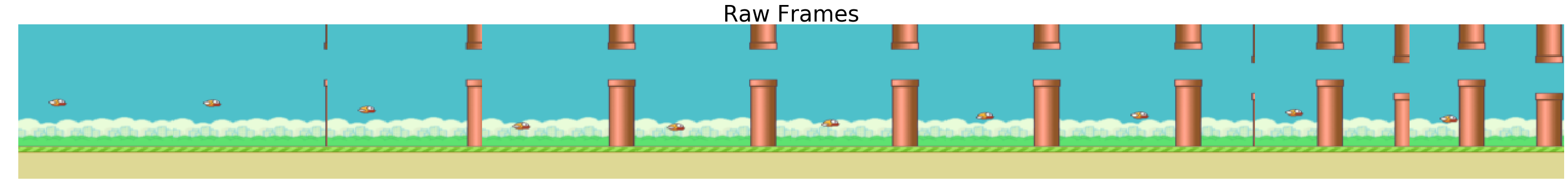}
\includegraphics[width=0.95\linewidth]{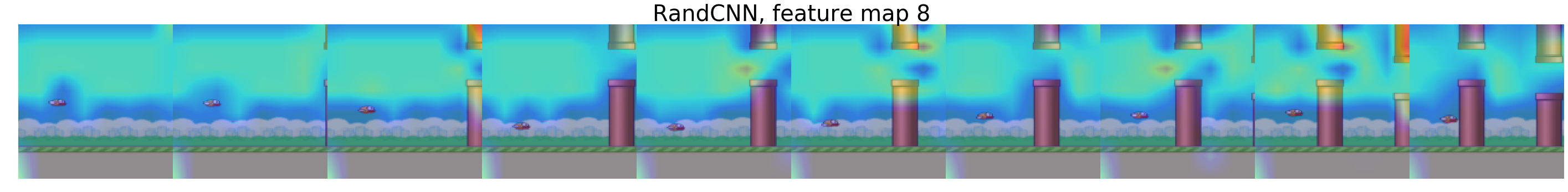}
\includegraphics[width=0.95\linewidth]{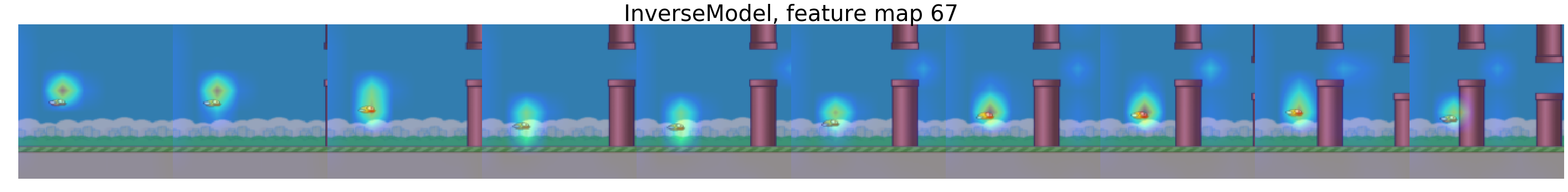}
\includegraphics[width=0.95\linewidth]{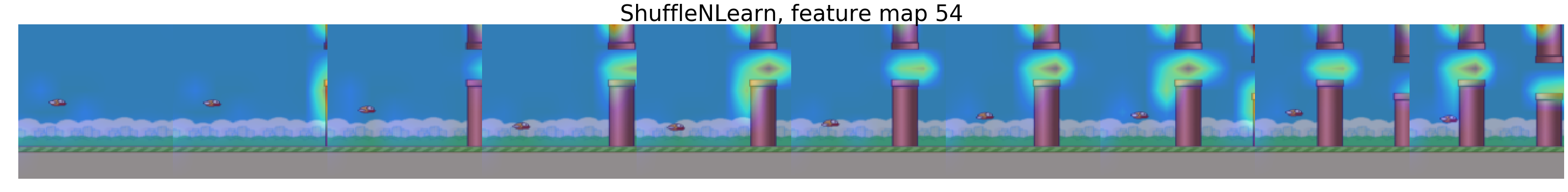}
\includegraphics[width=0.95\linewidth]{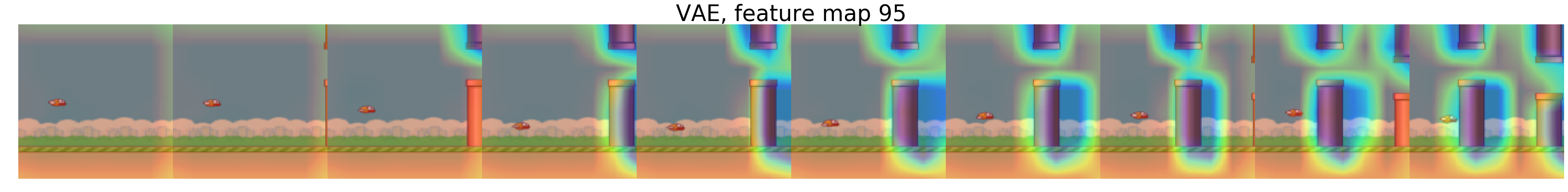}
\includegraphics[width=0.95\linewidth]{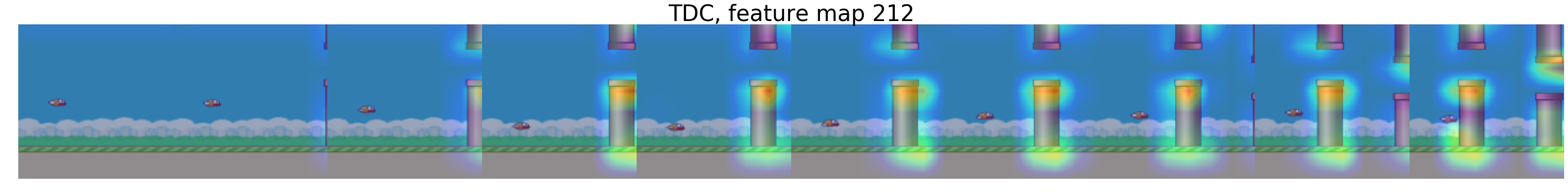}

\caption{\textbf{Qualititative Inspection of Feature Maps} (longer version) Flappy Bird feature maps from the last conv layer of the encoder superimposed on top of a sequence of frames they are a function of. Red pixels are high values, blue are low values}
\label{flappy-fmaps}
\end{center}
\vskip -0.2in
\end{figure*}
\end{document}